\documentclass[twoside,11pt]{article}

\usepackage[specialissue,accepted]{melba}

\usepackage{mwe} % to get dummy images, only for the example
\usepackage{amsmath,amsfonts}
\usepackage{multirow}
\usepackage{xcolor}
\usepackage{float}

\melbaid{2024:024}  % This is provided upon by the publishing editor
\doi{https://doi.org/10.59275/j.melba.2024-4862}
\melbaauthors{Atad et al.}  % Note: this one is also used to set the pdf 'authors' metadata
\volume{2}
\firstpageno{2103}  % Communicated by the publishing editor
\melbayear{2024}  % The publication year
\datesubmitted{03/2024}  % Date submitted to MELBA: mm/yyyy
\datepublished{09/2024}  % Today's date: mm/yyyy

\melbaspecialissue{Interpretability of Machine Intelligence in Medical Image Computing\\(iMIMIC) 2023}
\melbaspecialissueeditors{Mauricio Reyes, Jaime Cardoso, Jayashree Kalpathy-Cramer, Nguyen Le Minh, Pedro Abreu, José Amorim, Wilson Silva, Mara Graziani, Amith Kamath}

\ShortHeadings{Counterfactual Explanations for Medical Images using Diffusion Autoencoder}{Atad et al.}

\title{Counterfactual Explanations for Medical Image Classification and Regression using Diffusion Autoencoder}

\author{\name Matan Atad$^{1,2}$\orcid{0000-0001-6952-517X} \email matan.atad@tum.de
        \AND
        \name David Schinz$^{1}$\orcid{0000-0003-3734-1135}
        \AND
        \name Hendrik Moeller$^{1,2}$\orcid{0009-0001-1978-5894}
        \AND
        \name Robert Graf$^{1,2}$\orcid{0000-0001-6656-3680}
        \AND
        \name Benedikt Wiestler$^{1,4}$\orcid{0000-0002-2963-7772}
        \AND
        \name Daniel Rueckert$^{2}$\orcid{0000-0002-5683-5889}
        \AND
        \name Nassir Navab$^{3}$\orcid{0000-0002-6032-5611}
        \AND
        \name Jan S. Kirschke$^{1}$\orcid{0000-0002-7557-0003}
        \AND
        \name Matthias Keicher$^{3}$\orcid{0000-0003-2037-6796}
	\AND
        \addr 1. Department of Diagnostic and Interventional Neuroradiology, Klinikum rechts der Isar, Technical University of Munich, Germany\\
        \addr 2. Institute for Artificial Intelligence and Computer Science in Medicine, Technical University of Munich, Germany\\
        \addr 3. Computer Aided Medical Procedures, Technical University of Munich, Germany\\
        \addr 4. AI for Image-Guided Diagnosis and Therapy, Technical University of Munich, Germany
}

\begin{document}

% top matter
\maketitle

% abstract
\begin{abstract}%   <- trailing '%' for backward compatibility of .sty file
Counterfactual explanations (CEs) aim to enhance the interpretability of machine learning models by illustrating how alterations in input features would affect the resulting predictions. Common CE approaches require an additional model and are typically constrained to binary counterfactuals. In contrast, we propose a novel method that operates directly on the latent space of a generative model, specifically a Diffusion Autoencoder (DAE). This approach offers inherent interpretability by enabling the generation of CEs and the continuous visualization of the model’s internal representation across decision boundaries.

Our method leverages the DAE’s ability to encode images into a semantically rich latent space in an unsupervised manner, eliminating the need for labeled data or separate feature extraction models. We show that these latent representations are helpful for medical condition classification and the ordinal regression of severity pathologies, such as vertebral compression fractures (VCF) and diabetic retinopathy (DR). Beyond binary CEs, our method supports the visualization of ordinal CEs using a linear model, providing deeper insights into the model’s decision-making process and enhancing interpretability.

Experiments across various medical imaging datasets demonstrate the method’s advantages in interpretability and versatility. The linear manifold of the DAE’s latent space allows for meaningful interpolation and manipulation, making it a powerful tool for exploring medical image properties. Our code is available at ~\url{https://doi.org/10.5281/zenodo.13859266}.
\end{abstract}

% keywords

%%%%%%%%%%%%%%%%%%%%%%%%%%%%%%%%%%%%%%%%%%%%%%%%%%%%%%%%%%%%%%%%%%%%%%%%%%%
% Introduction
%%%%%%%%%%%%%%%%%%%%%%%%%%%%%%%%%%%%%%%%%%%%%%%%%%%%%%%%%%%%%%%%%%%%%%%%%%%
\begin{figure}[t]
  \centering
  \includegraphics[width=1.0\linewidth]{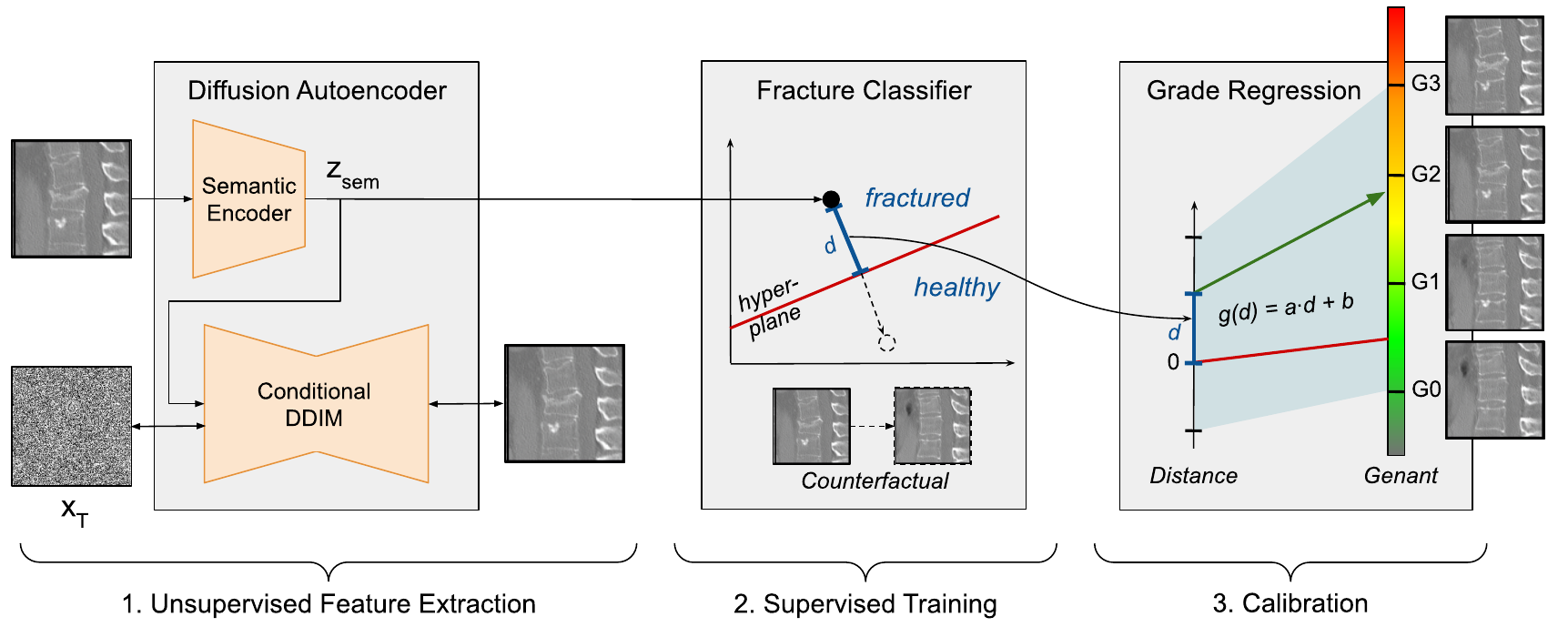}
  \caption{The proposed method involves three steps: 1) unsupervised training of a generative feature extractor Diffusion Autoencoder (DAE), 2) supervised training of a binary classifier to detect a pathology and obtain a decision hyperplane, and 3) calibrating a linear regression of the pathology grade to the hyperplane distance of embedded images. The method inherently enables the generation of counterfactual explanations (CEs), visualizing the model's representation corresponding to regression grades and smooth progressions in between.}
  \label{fig:graphical_abstract}
\end{figure}

\section{Introduction}

Generative models have emerged as a powerful tool for enhancing image classifiers' training process and interpretability. By synthesizing new images, these models can be employed for data augmentation, particularly in scenarios where certain classes are underrepresented. This oversampling technique has been shown to improve the performance of classifiers in various domains \citep{frid2018synthetic,sankar2021glowin}. Moreover, generative models can aid in the interpretability of classifiers by producing counterfactual explanations (CEs) \citep{wachter2017counterfactual}. These CEs illustrate the necessary realistic changes in an image that would alter the classifier's decision, providing valuable insights into the model's decision-making process.

However, existing methods for generating CEs often rely on external models that operate independently of the classifier, introducing a semantic gap and significant complexity \citep{atad2022chexplaining,bedel2023dreamr,pegios2024diffusion,fontanella2023diffusion}. In this paper, we propose an approach that directly operates on the latent space of a single model, specifically a Diffusion Autoencoder (DAE) \citep{preechakul2022diffusion}. Our method involves training the generative model to learn the data distribution in an unsupervised manner and then performing classification and regression tasks directly on the learned latent space.

By operating on the latent space of the DAE, our approach enables the inherent generation of CEs. These CEs provide a visual representation of the model's decision boundaries and illustrate the characteristics of images on either side of this boundary. This is particularly useful for modeling a regression between two extremes of a pathology, such as vertebral compression fractures (VCFs). In this context, obtaining labels for the intermediate grades between visibly distinct extremes is challenging due to label scarcity and inter-rater-disagreement. Our method addresses this issue by leveraging the DAE's continuous latent space to capture the pathology's progression. We interpolate between latent representations of extreme cases to perform ordinal regression in this latent space, modeling the continuous progression of the pathology. Moreover, since the generative model is trained without supervision and since our method is not dependent on an external classifier, we can generate CEs matching multiple tasks on the same latent space without repeated training.
Our contribution is three-fold:
\begin{enumerate}
    \item \textbf{Inherent CE generation}: The generative nature of DAE allows us to create images that resemble the model's inner representation of different pathology grades. With this approach, we extend binary CEs to ordinal regression and aid interpretability by visualizing the change in the image parts that the model deems most important. 
    \item \textbf{Unsupervised feature extraction}: We use unlabelled data to train a DAE model as an unsupervised feature extractor, this enables CE generation for multiple downstream tasks with the same learned latent space. Unlike previous methods that rely on separate models, our approach directly manipulates codes in the latent space, simplifying the process.
    \item \textbf{Modelling pathology grading as a continuous regression}: We employ a continuous regression approach in the latent space. We exemplify the strengths of this approach in modeling the smooth progression of pathologies such as intervertebral disc (IVD) degradation and diabetic retinopathy (DR) while requiring only binary labels.
\end{enumerate}

%%%%%%%%%%%%%%%%%%%%%%%%%%%%%%%%%%%%%%%%%%%%%%%%%%%%%%%%%%%%%%%%%%%%%%%%%%%
% Related works
%%%%%%%%%%%%%%%%%%%%%%%%%%%%%%%%%%%%%%%%%%%%%%%%%%%%%%%%%%%%%%%%%%%%%%%%%%%
\section{Related works}

\subsection{Counterfactual explanations (CEs)}

Counterfactual explanations (CEs), introduced by \citep{wachter2017counterfactual}, offer actionable insights by suggesting minimal changes to an input that would alter a model's classification. Unlike gradient-based explainability methods such as saliency maps \citep{wang2017chestx}, which highlight influential features, CEs aim for changes that maintain realism and adherence to the data distribution, distinguishing them from adversarial attacks \citep{verma2020counterfactual}.

In medical imaging, CEs are invaluable for providing clinicians with insights into alternative diagnoses through plausible image modifications. This capability is critical for enhancing trust in diagnostic models by ensuring CEs remain within the medical data distribution. Generation of medical imaging CEs has evolved with the application of generative models like Variational Autoencoders (VAEs) \citep{cohen2021gifsplanation} and Generative Adversarial Networks (GANs) \citep{singla2023explaining,atad2022chexplaining,schutte2021using, dravid2022medxgan}. Recent advances in controlled image generation with Diffusion Models \citep{dhariwal2021diffusion,ho2022classifier} have introduced sophisticated methods for creating condition-specific CEs. Techniques including classifier-guided \citep{bedel2023dreamr, pegios2024diffusion} leverage gradients from pretrained classifier and ones using classifier-free guidance \citep{sanchez2022healthy,dhinagar2024counterfactual,fontanella2023diffusion} condition the generation process on labels or classifier saliency maps to produce CEs.

Our method stands out by utilizing a Diffusion Autoencoder (DAE) \citep{preechakul2022diffusion} for CE generation within the diffusion model's latent space. This approach enables direct edits to latents, streamlining CE without needing external classifiers, and simplifying the process.

\subsection{Generative self-supervision}

Supervised Learning approaches have successfully solved many problems in image processing. Nevertheless, their dependency on the size of the annotated dataset is a significant obstacle in domains in which data annotation is time-consuming and relies on expert annotators. In Self-Supervised Learning (SSL), networks use unlabeled training data to learn meaningful feature representations through an auxiliary task, which are then used for downstream tasks. 

Specifically in the generative SSL setting, it is common to synthesize novel examples with a generative model and use them as a dataset for training an external model for a downstream task, such as classification \citep{frid2018synthetic,kitchen2017deep,nie2017medical}. Another line of work uses feature representations learned by the generative network itself for a discriminative task \citep{yi2018unsupervised,sankar2021glowin}. 

\cite{xu2021generative} demonstrated that latents learned by StyleGAN \citep{karras2020analyzing} can be used for downstream tasks such as regression and classification in a fully supervised manner. As a direct inspiration to our work, \cite{nitzan2021large} use StyleGAN latent codes to predict the magnitude of a visual attribute by measuring its distance from a hyperplane matching a linear direction with a semantic meaning. \cite{preechakul2022diffusion} migrated this approach to Diffusion Models, thus circumventing GAN Inversion \citep{tov2021designing}.

\subsection{Ordinal regression in clinical applications}

Medical image analysis requires distinguishing between binary classification, multiclass classification, ordinal regression, and regression. Binary classification differentiates between two distinct classes, such as the presence or absence of a pathology. Multiclass classification assigns cases to one of several categories without considering any order among them. In contrast, ordinal regression predicts a rank-ordered score that reflects the severity of a condition. Regression, on the other hand, predicts a continuous numerical value rather than discrete, ordered categories. While regression is used for predicting exact measurements, ordinal regression is used when the outcome is categorical but ordered, such as grading the severity of a disease. This ordered grading is crucial since it enables clinicians to monitor disease progression, adjust treatment plans, and prioritize cases based on severity (e.g., for DR \citep{flaxel2020diabetic}). Although this study focuses on ordinal regression, our method could also be applied to regression tasks where predicting a continuous outcome is required.

\subsubsection{Vertebral compression fractures (VCFs)} 

We specifically study VCFs, the most prevalent osteoporotic fractures in those over 50, causing significant pain and disability \citep{ballane2017worldwide,old2004vertebral}. Radiologists use the Genant scale \citep{genant1993vertebral} to measure fracture severity from CT images. Deep Learning can automate VCF detection 
\citep{valentinitsch2019opportunistic,tomita2018deep,chettrit2020,husseini2020grading,yilmaz2021automated,engstler2022interpretable,windsor2022context,iyer2023vertebral,hempe2024shape}, however, only a few works have considered VCF grading, all fully-supervised  \citep{pisov2020keypoints,zakharov2022interpretable,wei2022faint,yilmaz2023towards}. Compared to fracture detection, grading is an even more imbalanced task since medium to severely fractured vertebrae account for only a small portion of overall data. 
Closest to our approach, \cite{husseini2020conditioned} and \cite{hempe2024shape} train auto-encoders for vertebra shape reconstruction and then use the learned latent codes for downstream fracture detection. 

%%%%%%%%%%%%%%%%%%%%%%%%%%%%%%%%%%%%%%%%%%%%%%%%%%%%%%%%%%%%%%%%%%%%%%%%%%%
% Method
%%%%%%%%%%%%%%%%%%%%%%%%%%%%%%%%%%%%%%%%%%%%%%%%%%%%%%%%%%%%%%%%%%%%%%%%%%%
\section{Method}

Our method involves three steps (Fig.~\ref{fig:graphical_abstract}). First, the generative DAE model is trained using unlabelled data. This model learns a compressed, semantically rich, latent space useful for downstream tasks. Next, labeled images are encoded into this latent space, and a linear classifier is trained on these labels. Finally, we use the classifier's decision boundary to interpolate in latent space, edit images, create CEs, and rate pathologies.

\subsection{Unsupervised feature extraction}

We use DAE \citep{preechakul2022diffusion} as a generative unsupervised feature extractor. DAE leverages the capabilities of Denoising Diffusion Probabilistic Models (DDPMs) \citep{sohl2015deep} and Denoising Diffusion Implicit Models (DDIMs) \citep{song2020denoising} for unsupervised feature extraction. DAE distinguishes itself by incorporating a semantic encoder, designed to transform input images into a semantic latent space $z_{\text{sem}}$, that captures semantically meaningful information. This space is different from the DDPM and DDIM noise latent space $x_{\text{T}}$, which was shown to lack high-level semantics \citep{kwon2022diffusion}. Using $z_{\text{sem}}$ enables the model to generate high-fidelity reconstructions and facilitate downstream tasks with a higher degree of semantic awareness. Moreover, it was shown that the semantic latent space is characterized by a linear data manifold, similar to StyleGAN's StyleSpace \citep{preechakul2022diffusion}, facilitating meaningful interpolation between latent representations.

The architecture of DAE is characterized by its use of conditional DDIM, which performs dual functions: it acts as a stochastic encoder to encode the input image into a noise representation $x_{\text{T}}$ and as a decoder to reconstruct the image from the combined semantic and noise latents. This method allows the model to effectively retain and utilize semantic details throughout image reconstruction. Training the DAE model involves minimizing the loss function:
\begin{align}
L = \mathbb{E}_{x_0, t}\left[ ||\theta(x_t, t, z_{\text{sem}}) - \epsilon_t||^2 \right]
\end{align}
where $x_0$ is the input image, $t$ is the timestep, $\epsilon_t$ is the noise component and $\theta$ is a neural network.  Training the DDIM backbone and the semantic encoder simultaneously ensures that the model produces semantically meaningful and visually coherent outputs.

\subsection{Linear decision boundary}

We obtain the semantic latent representation $z_{sem}$ for a subset of training samples for which labels are available and train linear classifiers (linear regression and SVM) to predict the existence of the target pathology. The decision boundary of a binary linear classifier is represented by the hyperplane given by its normal equation:
\begin{align}
\vec{P}: \vec{n} \cdot \vec{w} + b
\end{align}
where $\vec{n}$ is a semantic direction corresponding to the pathology existence, $\vec{w}$ is an input image latent, and $b$ is a bias term. The magnitude of pathology in grading tasks is estimated using the distance of the latent $\vec{w}$ to the hyperplane \citep{nitzan2021large}: 
\begin{equation}
\text{dist}(\vec{w}, \vec{P}) = \frac{\vec{n} \cdot \vec{w} + b}{\|\vec{n}\|}
\end{equation}
A simple linear regression is fitted to calibrate this distance to the respective pathology scale. Finally, the continuous values are rounded to the nearest grade to obtain the clinical grading in ordinal categories, matching the ground truths.

\subsection{Counterfactual explanations}
To generate CE images, the semantic latent code can be changed in the direction $\vec{n}$ and together with the original stochastic latent decoded by the conditional DDIM to a new image. For binary CEs, we reflect a given latent sample to an equivalent position on the opposite side of the decision boundary, maintaining the original distance from the boundary within the latent space. For a latent sample $\vec{w}$, the counterfactual is:
\begin{align}
\vec{w}_{\text{ce}} = \vec{w}- 2 \cdot \text{dist}(\vec{w}, \vec{P})
\end{align}

To generate a counterfactual of a particular pathology grade, the calibration utilized in the regression process is inverted to determine the magnitude of the required change along the semantic direction. 

%%%%%%%%%%%%%%%%%%%%%%%%%%%%%%%%%%%%%%%%%%%%%%%%%%%%%%%%%%%%%%%%%%%%%%%%%%%
% Experiments
%%%%%%%%%%%%%%%%%%%%%%%%%%%%%%%%%%%%%%%%%%%%%%%%%%%%%%%%%%%%%%%%%%%%%%%%%%%
\section{Experiments}

\subsection{Datasets} 

\begin{table}[htbp]
  \centering
  \caption{Distribution of labeled samples across datasets used in this study.}
  \resizebox{\textwidth}{!}{
  \begin{tabular}{c|c|c|c|c|c|c}
    Dataset & Task & label 0 & label 1 & label 2 & label 3 & label 4 \\ 
    \hline
    VerSe & Genant grading & $4124$ & $74$ & $102$ & $44$ & - \\
    \hline
    SPIDER & Pfirrmann grading & $217$ & $339$ & $417$ & $291$ & $182$ \\
    \hline
    RetinaMNIST & Diabetic retinopathy severity & $540$ & $140$ & $234$ & $214$ & $72$ \\
    \hline
    BraTS & Peritumoral edema & $203$ & $281$ & - & - & - \\
    \hline
    \multirow{4}{*}{MIMIC-CXR} & Lung opacity & $66130$ & $28465$ & - & - & - \\
    & Pneumonia & $66096$ & $2629$ & - & - & - \\
    & Cardiomegaly & $66094$ & $25201$ & - & - & - \\
    & Edema severity grading & $2908$ & $1631$ & $1873$ & $725$ & - \\
    \hline
  \end{tabular}
  }
  \label{tab:dataset_size}
\end{table}

We experimented with several ordinal regression and classification tasks, some suffering from high class-imbalance (Table~\ref{tab:dataset_size}).

\noindent \textbf{Vertebral compression fractures (VCFs):} We used the public VerSe dataset \citep{liebl2021computed,sekuboyina2021verse} and an in-house dataset acquired at Klinikum Rechts der Isar and Klinikum der Universität München \citep{foreman2024deep}, containing a total of 12019 sagittal 2D CT slices of vertebrae. Each slice has a size of $96 \times 96$ pixels centered around a single vertebra, though multiple surrounding vertebrae are also visible. For each slice, the existence of a VCF in the center vertebra is indicated, and a small subset also includes a Genant grading, ranging from G0 (Normal/No fracture) to G3 (Severe/Over 40\% reduction). Of the 1248 fractured samples, 220 have a grading (74 G1, 102 G2, 44 G3). We applied no image augmentations and kept other training parameters with their defaults. We only use samples from the VerSe dataset to train the fracture classifiers and evaluate their grading.

\noindent \textbf{Intervertebral disc (IVD) degeneration:} The SPIDER dataset \citep{van2024lumbar} contains a total of 1446 3D T2-weighted MRI volumes. We preprocess these volumes to extract a random 2D $64 \times 64$ sagittal slice, each centered around a single IVD. Each IVD's matching Pfirrmann grade \citep{pfirrmann2001magnetic} is indicated. The Pfirrmann grading system is a widely recognized scale for assessing the degree of IVD degeneration in MRI. It ranges from Grade I, indicating no degeneration with a homogeneous structure and normal disc height, to Grade V, which signifies severe degeneration with a collapsed disc space and inhomogeneous structure with areas of hypointensity. 

\noindent \textbf{Diabetic retinopathy (DR):} The RetinaMNIST dataset \citep{medmnistv2} comprises 1200 $128 \times 128$ pixel Fundus camera images, each labeled with its corresponding DR severity grade. DR is a diabetes complication that affects the eyes and is characterized by damage to the blood vessels of the light-sensitive tissue at the back of the eye (retina). The severity of DR is categorized into five stages: from Grade 0 (no DR), representing no apparent retinal damage, to Grade 4 (proliferative DR), indicating advanced disease with a high risk of vision loss. Due to the small size of the dataset, we further applied image augmentation while training the DAE: image rotation ($30^{\circ}$), image flip, grid distortion, and zoom ($90\%-110\%$), each with a probability of $0.5$.

\noindent \textbf{Peritumoral edema in brain tumors:} The BraTS dataset \citep{menze2014multimodal} consists of 484 3D FLAIR T2-weighted MRI volumes. In the preprocessing step, we extract a random 2D $64 \times 64$ slice from each volume and a binary label indicating the presence of peritumoral edema in each slice. This condition represents fluid accumulation around the tumor site, often indicative of the tumor's aggressiveness and the body's response to its presence.

\noindent \textbf{Cardiomegaly, lung opacity, pneumonia and edema in chest X-rays:} The MIMIC-CXR dataset \citep{johnson2019mimic} is an extensive collection of chest radiographs labeled to reflect common conditions that can be identified through radiographic imaging. In the preprocessing step, each chest X-ray image is resized to a $256 \times 256$ resolution to standardize input size across the dataset. The DAE was trained on 166512 unlabeled images. For training the classifiers, we chose a subset of labeled anterior-posterior (AP) images for lung edema, opacity, pneumonia, and cardiomegaly and used labels extracted from reports by the CheXbert~\citep{smit2020chexbert} labeler. The severity grades for the ordinal regression of edema are taken from \cite{horng2021deep} following their dataset splits.

\subsection{Implementation details}

We trained the DAE for $12,000,000$ steps with a batch size of $64$ on a single Nvidia A40 GPU. We used the official DAE implementation by \cite{preechakul2022diffusion}. The semantic encoder used codes of $1 \times 512$. To encode images and create CEs, $250$ reverse diffusion steps were used to encode the images back to the stochastic latent $x_T$. The stochastic latent was then concatenated to the semantic latent in $100$ forward diffusion steps to retrieve the original image $x_0$. More training hyper-parameters are given in Sec.~\ref{appendix}. All grades apart from the minimal are considered positive for training the SVM and linear regression classifiers. For VCFs, since the labels were noisy, only G2 and G3 were considered as the positive class. For the image reconstruction baseline, we used StyleGAN2 \citep{karras2020analyzing} with Encoder4Editing (E4E) \citep{tov2021designing}. For the classification and regression baseline, we used DenseNet121 and trained it from scratch with the same data pipeline.

\subsection{Evaluation metrics}

To evaluate image reconstruction, we measure the perceptual similarity between the original and encoded image using Learned Perceptual Image Patch Similarity (LPIPS) and the general image generation quality using the Fréchet Inception Distance (FID). For classification performance we use ROC-AUC and $F_1$ score. We calculated the macro $F_1$ and Mean Average Error (MAE) for grading performance.

\section{Results and discussion}\label{discussion}

\subsection{Image reconstruction}

\begin{figure}[htbp]
  \centering
  \includegraphics[width=0.6\linewidth]{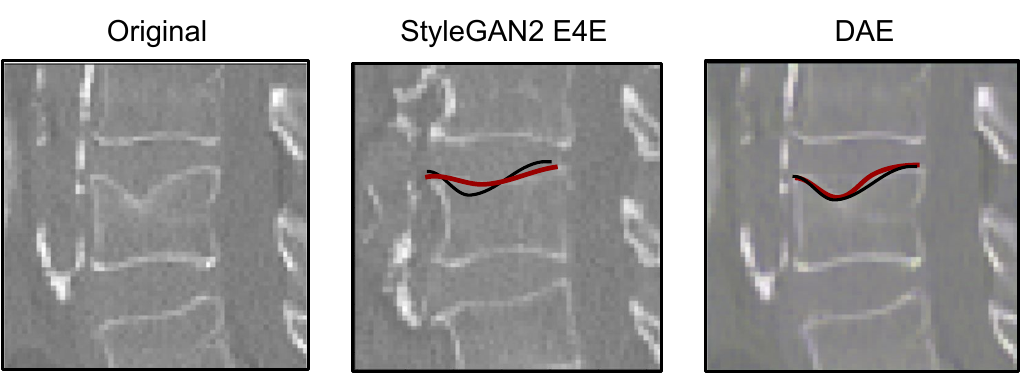}
  \caption{
Qualitative comparison of VCF encoders: The original shape of the evaluated vertebra is highlighted in black, while the reconstructed shape is shown in red. Compared to StyleGAN2 \citep{karras2020analyzing} with Encoder4Editing (E4E) \citep{tov2021designing}, the DAE shows the closest resemblance to the original.
  }
  \label{fig:compare_encoders}
\end{figure}

We begin by evaluating the DAE’s capability to accurately reconstruct images from latent codes using the VCF task. As shown in Table~\ref{tab:combined}, the DAE semantic encoder outperforms StyleGAN2 E4E in reconstructing vertebrae images, as evidenced by LPIPS scores. Fig.~\ref{fig:compare_encoders} supports this finding, illustrating E4E’s limited ability to capture fracture-relevant features. Despite the overall similarity in appearance between the original and E4E-reconstructed images, the fracture cavity (highlighted in red) is mostly lost. Additionally, the FID measurements in Table~\ref{tab:combined} demonstrate that DAE generates higher-quality image distributions.
\begin{table}[htbp]
  \centering
  \caption{Quantitative evaluation of VCF image encoding and generation. DAE shows the best performance in encoding and generation.}
\resizebox{0.8\textwidth}{!}{
  \begin{tabular}{l|l|l|l}
    Model & Encoder & Encoding (LPIPS $\downarrow$) & Generation (FID $\downarrow$) \\
    \hline
    StyleGAN2 & E4E & $0.098$ & $134$ \\
    DAE & DAE encoder & $\mathbf{0.040}$ & $\mathbf{40}$ \\
  \end{tabular}
  }
  \label{tab:combined}
\end{table}

\subsection{Discriminative tasks}

We conduct extensive experiments to measure the discriminative capability of the DAE latents in detection and regression in multiple tasks. For VCFs, Table~\ref{tab:fracture_results} shows that both linear layers and SVM can effectively separate fractured from healthy vertebrae by constructing a separating hyperplane with a respective AUC of 0.96 and 0.93. Results reported by \cite{husseini2020conditioned} further highlight the advantages of DAE as a feature extractor in comparison to Autoencoder (AE) and Variational Autoencoder (VAE). Although the linear layer achieves better detection results, the SVM’s hyperplane shows better performance in the linear regression of Genant grades. In Table~\ref{tab:other_detection_grading}, our method achieved comparable results to a fully supervised DenseNet121 baseline on other detection tasks. 
\begin{table}[htbp]
  \centering
    \caption{Quantitative evaluation for VCF detection and grading. The linear layer on top of DAE did not converge for grading due to severe data imbalance. Results marked with $^*$ use 3D volumes and have been evaluated on a different test split.}
     \resizebox{0.95\textwidth}{!}{
      \begin{tabular}{l|l|l|l|l}
        Model & Encoder & Classifier & Detection (AUC $\uparrow$) & Grading ($F_1$ $\uparrow$)  \\
        \hline
        \multicolumn{5}{c}{\textit{Linear probing trained on G0, G2 and G3 with frozen encoder of generative model}}\\
        \hline
        AE \citep{husseini2020conditioned} & AE &  Linear layer & $0.70^{\ast}$ & - \\
        VAE \citep{husseini2020conditioned} & VAE &  Linear layer & $0.77^{\ast}$ & - \\
        Spine-VAE \citep{husseini2020conditioned} & VAE &  Linear layer & $0.81^{\ast}$ & - \\
        \textbf{3D} point-cloud \citep{hempe2024shape} & AE & MLP & $0.93^\ast$ & - \\
        StyleGAN2 & E4E & SVM & $0.74$ & -\\
        
        DAE & DAE & Linear layer&  0.96 &  (0.23)\\
        DAE& DAE &  SVM & 0.93 &  0.59\\
        \hline
        \multicolumn{5}{c}{\textit{Linear regression with distance to the hyperplane, calibrated with means of G0 and G3}}\\
        \hline
        DAE & DAE & Linear layer&  0.96 & 0.44 \\
        DAE& DAE &  SVM & 0.93 &  0.51\\
        \hline
        \multicolumn{5}{c}{\textit{Polynomial regression with distance to hyperplane, calibrated with G0, G2 and G3}}\\
        \hline
        DAE& DAE &  SVM  (deg=1)& 0.93 &  0.42\\
        DAE& DAE &  SVM  (deg=3)& 0.93 &  0.56\\
        \hline
        \multicolumn{5}{c}{\textit{End-to-end training with full supervision  (G0, G2 and G3)}}\\
        \hline
        \multicolumn{3}{l|}{DenseNet121, baseline} & $0.98$ & $0.65$ \\
        \multicolumn{3}{l|}{\textbf{3D} SE-ResNet50 with SupCon loss \citep{wei2022faint}} & $0.99^\ast$ & $0.86^\ast$ \\
      \end{tabular}
  \label{tab:fracture_results}
  }
\end{table}
\begin{table}[htbp]
  \centering
  \caption{Quantitative evaluation for other detection and grading tasks. The proposed method presents comparable results to a fully-supervised DenseNet121.}
  \resizebox{0.95\textwidth}{!}{
  \begin{tabular}{c|c|c|c|c|c|c}
    Model & Dataset & Task (\# labels)& \multicolumn{2}{|c|}{Detection} & \multicolumn{2}{|c}{Grading} \\
    & & & AUC $\uparrow$ & $F_1$ $\uparrow$ & MAE $\downarrow$ & $F_1$ $\uparrow$ \\
    \hline
    DAE, SVM classifier & \multirow{3}{*}{Spider} & \multirow{3}{*}{Pfirrmann grade, ordinal (5)}  & $\mathbf{0.67}$ & $\mathbf{0.93}$ & $0.89$ & $0.25$\\
    DAE, LR classifier & & & $0.65$ & $\mathbf{0.93}$ & $0.87$ & $\mathbf{0.33}$\\
    DenseNet121 & & & $0.64$ & $0.92$ & $\mathbf{0.83}$ & $0.30$\\
    \hline
    DAE, SVM classifier & \multirow{3}{*}{RetinaMNIST} & \multirow{3}{*}{Diabetic retinopathy severity, ordinal (5)}  & $\mathbf{0.83}$ & $\mathbf{0.87}$ & $0.72$ & $0.31$\\
    DAE, LR classifier & & & $0.82$ & $0.86$ & $\mathbf{0.71}$ & $0.25$\\
    DenseNet121 & & & $0.81$ & $0.86$ & $0.75$ & $\mathbf{0.32}$\\
    \hline
    DAE, SVM classifier & \multirow{3}{*}{BraTS} & \multirow{3}{*}{Peritumoral edema, binary (2)}  & $0.57$ & $0.94$ & - & - \\
    DAE, LR classifier & & & $\mathbf{0.63}$ & $0.94$ & - & -\\
    DenseNet121 & & & $0.49$ & $\mathbf{0.96}$ & - & -\\
    \hline
    DAE, SVM classifier & \multirow{2}{*}{MIMIC-CXR}  & \multirow{2}{*}{Edema severity grade, ordinal (4)} &  $0.68$ & $0.76$ & $0.77$ & $0.24$ \\
    DAE, LR classifier &  &  & $\mathbf{0.69}$ & $0.76$ & - & - \\
    \hline
  \end{tabular}
  \label{tab:other_detection_grading}
  }
\end{table}

\begin{figure}[h]
\centering
\includegraphics[width=0.8\linewidth]{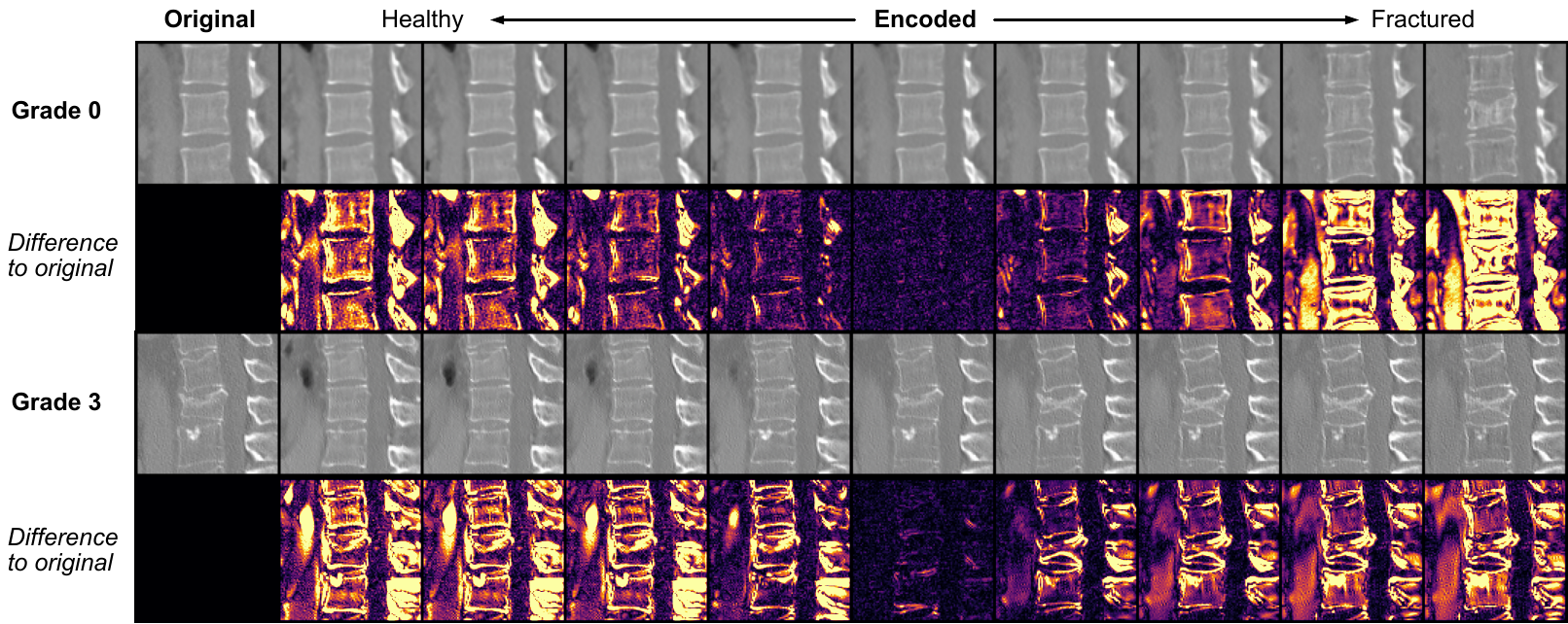}
  \caption{Images generated by moving the semantic latent orthogonal to the hyperplane without calibration. Top row: healthy vertebra (G0) moved in both directions, revealing a severe fracture on the right. Bottom row: severely fractured vertebra (G3) decompresses on the left and further disintegrates on the right. A hallucination of a lung is added by the model to both images when semantically shifted further into the healthy direction.}
  \label{fig:edited_images_shift}
\end{figure}

\begin{figure}[htbp]
  \centering
  \includegraphics[width=0.67\linewidth]{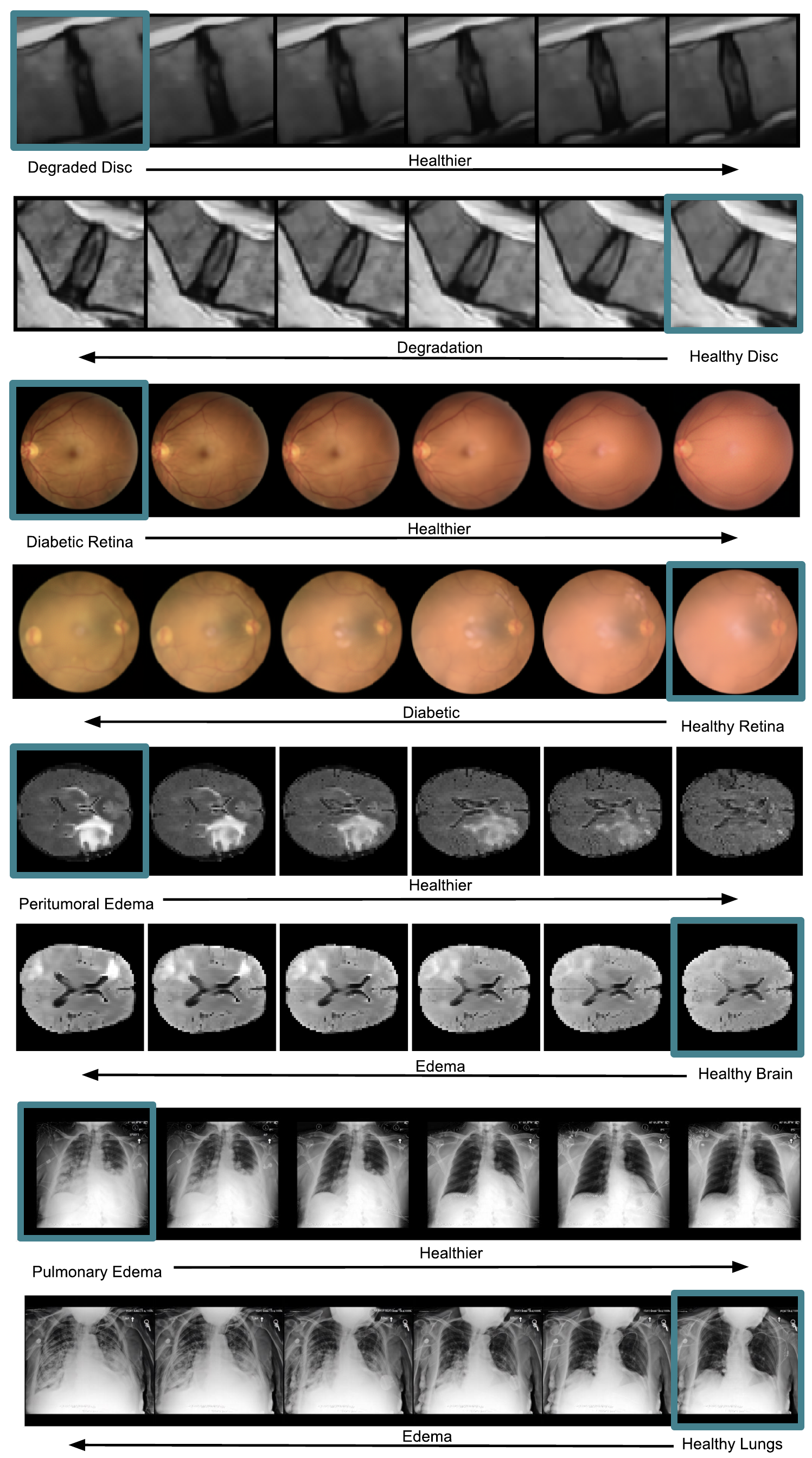}
  \caption{Interpolation in the latent space. In each row, the original image is in the blue box. The rest show the progression of the pathology, edited by moving the image latent perpendicularly to a hyperplane of a binary classifier in the DAE semantic latent space.}
  \label{fig:image_editing_figure}
\end{figure}

\subsection{Counterfactual explanations}

Besides feature extraction, our method provides inherent CE generation. In this setting, the generative capability of the model can aid the interpretability of the decision support system. In Fig.~\ref{fig:edited_images_shift}, the input image (left), after being encoded into the latent space (center), can be manipulated towards both classes, e.g., from healthy vertebra to severe fracture and vice versa. Similarly, in Fig.~\ref{fig:image_editing_figure}, the input image (in the blue box) is manipulated towards the opposite class relative to the binary decision boundary, i.e., from healthy to pathology and vice versa.

\begin{figure}
\includegraphics[width=0.95\linewidth]{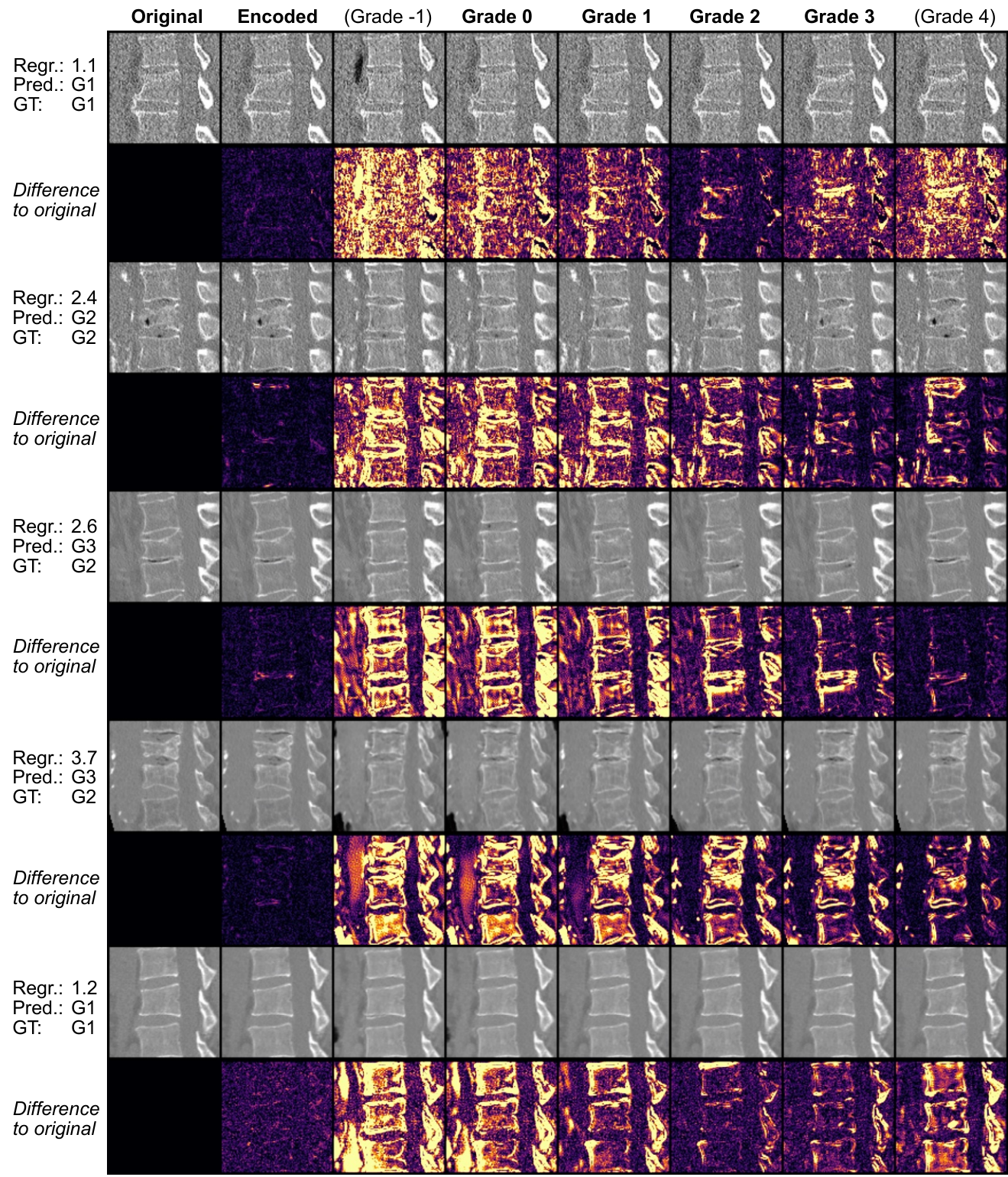}
  \caption{DAE image generation calibrated to Genant grades (linear regression to SVM hyperplane): On the left, the results of regression, prediction, and the ground truth (GT) are shown. The first three rows are well-calibrated examples, while the bottom two rows show examples that are not well-calibrated. Note that G1 was not used for training the classifiers, and neither G1 nor G2 was used for calibrating the regressors. The artificial scores -1 and 4 are added for illustrational purposes only.}
  \label{fig:edited_images_calibrated}
\end{figure}

\begin{figure}[tbp]
  \centering
  \includegraphics[width=1.0\linewidth]{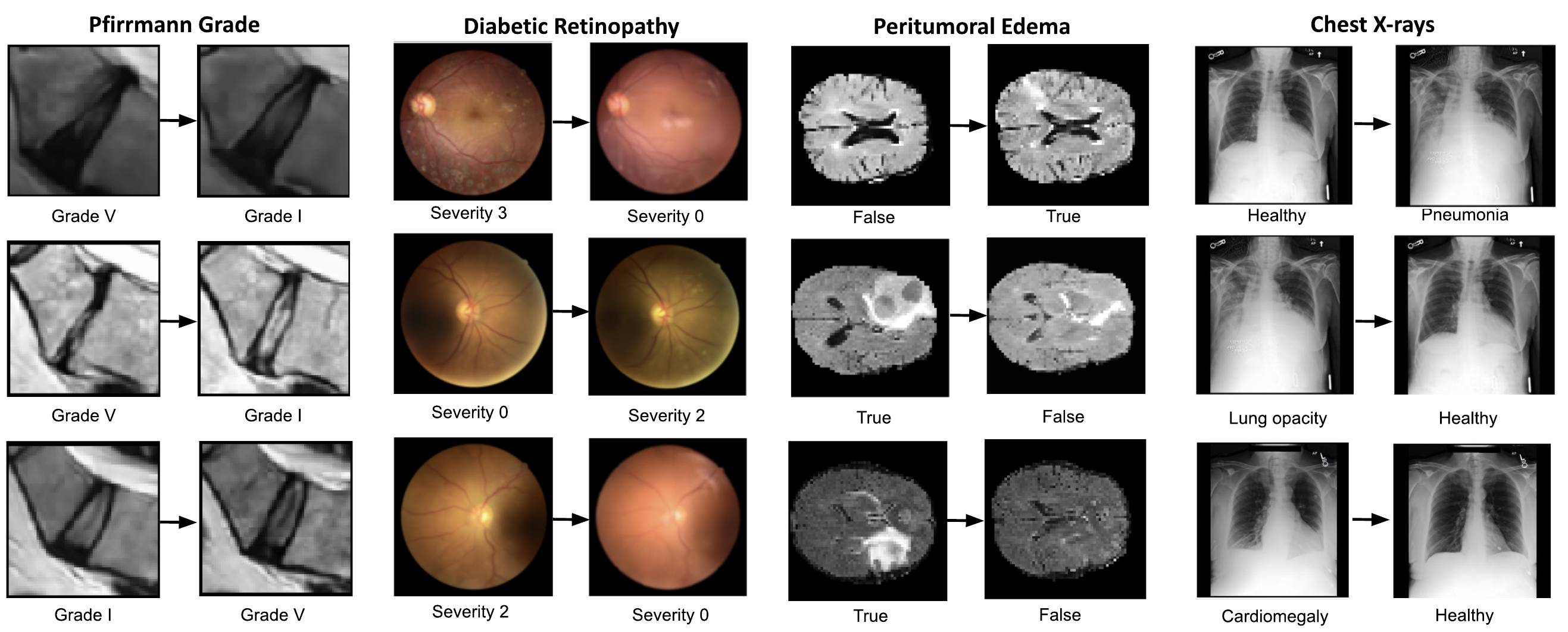}
  \caption{Regression and binary CEs generated using our method: In each pair of images, the left is the original with its matching ground-truth label, and the right is the CE with its predicted classifier label or regressor score.}
  \label{fig:regression_cunterfactuals}
\end{figure}

Fig.~\ref{fig:edited_images_calibrated} demonstrates that our CEs can be extended to visualize whether the model is well calibrated to the ordinal regression of Genant grades by showing the model's perception of each grade for a given vertebra. While the last row in Fig.~\ref{fig:edited_images_calibrated} is not well calibrated, the rightmost image augments a barely visible feature. This exaggeration of anatomical changes could guide the radiologist's attention and help estimate the potential progression of pathological changes barely visible in the original image.

Additional examples in Fig.~\ref{fig:regression_cunterfactuals} highlight the limitations of our method. A qualitative evaluation by a clinician revealed several issues: The Pfirrmann grade V CE shows only a slight signal loss with minimal height loss, suggesting it should be graded as grade II. In the DR examples, the model introduces additional blurring when transitioning from diseased to healthy states. The removal of peritumoral edema is only partial in the second case. Additionally, in the same task, clinicians noted size changes between the CEs and the original images, indicating a potential model bias related to the MRI slice position along the axial plane.

\subsection{Limitations and future work} 

\subsubsection{Discriminative tasks}
While the well-calibrated samples in Fig.~\ref{fig:edited_images_calibrated} show impressive visual results, the VCF grading metrics in Table~\ref{tab:fracture_results} reveal the limitations of our method. The linear separability of classes in DAE's semantic latent space using 2D slices cannot compete with the fully supervised, end-to-end baseline and 3D methods, motivating an extension to three dimensions. 

One reason for this might be a failure to disentangle the fracture of adjacent vertebrae from the central vertebral body that is classified. This can be observed in the second to last example in Fig.~\ref{fig:edited_images_calibrated}, where the center vertebra remains unchanged while the top one changes with the severity of the fracture. At the same time, this showcases the interpretability of our approach by visualizing the model's understanding of different grades, highlighting its misguided feature attribution. In future work, the network could be explicitly guided to attend to the central vertebrae, or the generation process could be conditioned on anatomical features, thus disentangling these from disease progression.

\begin{figure}[tbp]
\includegraphics[width=1\linewidth]{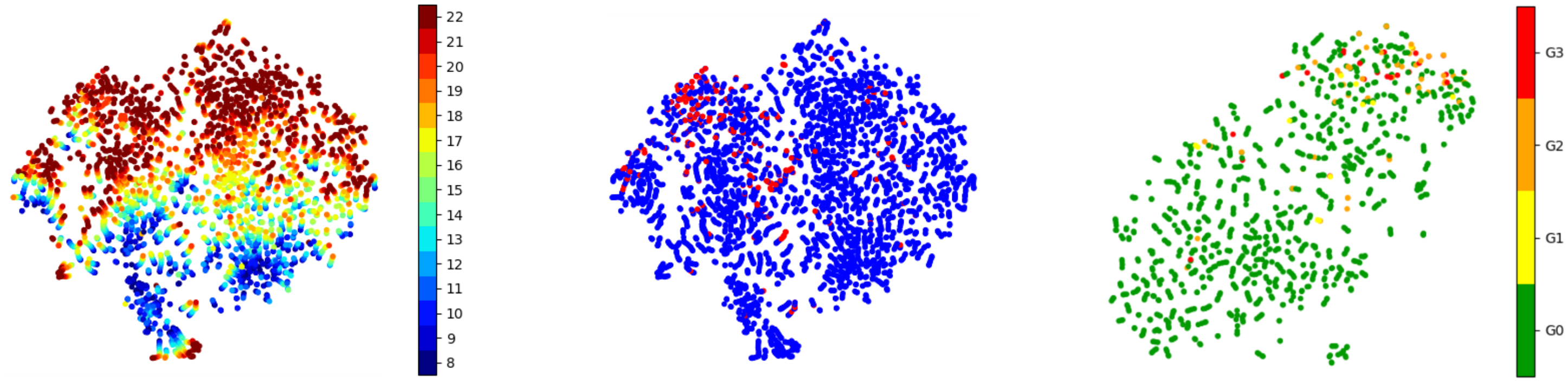}
  \caption{t-SNE projections of $z_{sem}$ of VerSe training set vertebrae encoded by the DAE model. On the left, the vertebra levels from T1 (8) to L5 (22) are visualized. In the center, the same data with healthy vertebrae (blue) and fractures (red) is indicated. On the right, the levels L1-L4 are shown with their Genant grades ranging from G0 to G3}
  \label{fig:tsne}
\end{figure}

The t-SNE projection of latents in Fig.~\ref{fig:tsne} clearly shows that the unsupervised DAE clustered the vertebral levels ranging from T1 to L5. Since fractures occur more frequently in the lumbar spine, a cluster of fractures can be observed there with many outliers. 
We hypothesized that training with all levels could aid the data imbalance since only very few fractures are present in the data per vertebral level. However, the better separatability of Genant grading in the visually similar subset of vertebra L1-L4 (Fig.~\ref{fig:tsne}, right) suggests investigating independent models for the different spine segments. 

To challenge the assumption of a linear relationship between the distance to the hyperplane and the severity of the fracture, we fitted a polynomial regression to grades G0, G2, and G3. The results in Table \ref{tab:fracture_results} show an improvement over our simple regression model, indicating the presence of a non-linear relationship that warrants further investigation in future research.

\subsubsection{Counterfactual explanations}
This work has demonstrated how CEs can reveal model biases by uncovering the inner representations of different classes and regression scores. For instance, manipulating vertebral images towards the healthy direction introduced a lung artifact, effectively translocating the vertebrae upwards within the spine where fractures are uncommon (Fig.~\ref{fig:edited_images_shift}). Another example is the bias related to brain slice position along the axial plane and the presence of edema (Fig.~\ref{fig:image_editing_figure}). 

Beyond revealing such model biases, CEs offer crucial insights into AI models, lifting the "black box" and empowering clinicians (and researchers) to understand the models' inner decision-making. Further, CEs might potentially drive the discovery of novel imaging biomarkers \citep{kumar2022counterfactual}. 

Future studies should validate the clinical applications of CEs based on DAE. These studies should focus on several key areas. First, they should explore the potential of CEs for imaging biomarker discovery by unveiling which features drive classification. Second, researchers should assess whether generated CEs can effectively demonstrate potential disease progression, helping clinicians anticipate future developments. Third, studies could investigate using CEs to visualize reverse disease progression, e.g. aiding in planning reconstructive surgeries such as bone cement filling in vertebral bodies. Lastly, these investigations should aim to identify model biases, ultimately improving the overall reliability of the algorithm in real-world clinical settings. Through these diverse applications, CEs could provide valuable insights into disease mechanisms and treatment planning while enhancing the interpretability and trustworthiness of AI models in clinical practice.

\section{Conclusion}

We present a methodological framework for generating counterfactual explanations (CEs) in medical imaging using Diffusion Autoencoders (DAE). By directly manipulating the latent space, our approach avoids the need for external classifiers, simplifying the generation process. Our findings highlight the versatility and interpretability of the DAE model across various medical imaging tasks, proving effective in classification and regression.

Moreover, this work demonstrates how CEs can uncover model biases. Future work will focus on the clinical validation of CEs to assess their practical applications, such as the discovery of imaging biomarkers and guiding treatment decisions. Additionally, we aim to address the study's limitations, particularly in extending the model to three-dimensional data and disentangling the disease progression from anatomical features in the CE generation process.

%%%%%%%%%%%%%%%%%%%%%%%%%%%%%%%%%%%%%%%%%%%%%%%%%%%%%%%%%%%%%%%%%%%%%%%
% Mandatory Sections. Please complete, especially for final publication
%%%%%%%%%%%%%%%%%%%%%%%%%%%%%%%%%%%%%%%%%%%%%%%%%%%%%%%%%%%%%%%%%%%%%%%

% Acknowledgements.
% Please include any funding, intellectual contributions not included in the authorship, and any other acknowledgements.
\acks{The authors acknowledge the financial support by the Federal Ministry of Education and Research of Germany (BMBF) under project DIVA (FKZ 13GW0469C) and from the European Research Council (ERC) under the European Union’s Horizon 2020 research and innovation program (101045128—iBack-epic—ERC2021-COG).}

% Ethical Standards.
% Please edit with the appropriate ethics considerations for your work. Include any pertinent IRB information, etc.
%
% Please note that the submission requirements included:
% The work presented must follow appropriate ethical standards in conducting research and writing the manuscript, following all applicable laws and regulations regarding treatment of animals or human subjects.
\ethics{The work follows appropriate ethical standards in conducting research and writing the manuscript, following all applicable laws and regulations regarding treatment of animals or human subjects.}

% Conflict of Interest
% Declaration of possible conflicts of interest: Authors must disclose any financial, organisational, commercial or personal conflicts of interest that might bias their work.
% If no conflicts, please say "We declare we don't have conflicts of interest."
\coi{JK is co-founder of Bonescreen GmbH. The remaining authors declare the research was conducted in the absence of conflicts of interest.}

% Data availability
\data{The majority of data used for this study is publicly available on the respective challenge's website, allowing for reproduction of our results. We further used an internal dataset of unlabeled spine CTs for the VCF task, which could not be shared due to patient privacy.}

\bibliography{references}

% Manual newpage inserted to improve layout of sample file - not
% needed in general before appendices.
% \newpage

% Appendix is optional
\clearpage
\appendix
\section*{Appendix}\label{appendix}
\begin{table}[!htb]
\centering

\caption{Description of the in-house VCF dataset from Klinikum Rechts der Isar and Klinikum der Universität München used for training the DAE.}
\begin{tabular}{|l|l|}
\hline
\textbf{Characteristic}              & \textbf{Value}                                 \\ \hline
Number of patients                   & 465                                           \\ \hline
Median age (years)                   & $\sim$69 ($\pm$12)                             \\ \hline
Scan types                           & Healthy and fractured vertebrae               \\ \hline
Nature of fractures                  & Osteoporotic or malignant                     \\ \hline
Additional features                  & Metallic implants and foreign materials       \\ \hline
CT scanner                        & Heterogeneous              \\ \hline
Scanner setting                 & Heterogeneous              \\ \hline
Field of view & Varying \\ \hline
\end{tabular}
\label{table:dataset_summary}
\end{table}

\begin{table}[!htb]
\centering

\caption{Dependencies and their respective versions required for implementing the proposed method. The DAE and all neural networks were trained with PyTorch. The SVM classifier and logistic regression were trained with cuML.}
\begin{tabular}{|l|l|}
\hline
\textbf{Library}         & \textbf{Version} \\ \hline
python                    & 3.9.15           \\ \hline
torch                    & 1.8.1           \\ \hline
torchvision              & 0.9.1           \\ \hline
monai                    & 1.0.1           \\ \hline
pytorch-lightning        & 1.4.5           \\ \hline
torchmetrics             & 0.5.0           \\ \hline
scipy                    & 1.5.4           \\ \hline
numpy                    & 1.19.5          \\ \hline
pytorch-fid              & 0.2.0           \\ \hline
lpips                    & 0.1.4           \\ \hline
pandas                   & 1.1.5           \\ \hline
Pillow                   & 8.3.1           \\ \hline
lmdb                     & 1.2.1           \\ \hline
cuml                     &     22.10.1 \\ \hline
scikit-learn              &    1.1.3 \\ \hline
scikit-image               &   0.19.3 \\ \hline
\end{tabular}
\label{table:dependencies}
\end{table}

\begin{table}
\centering

\caption{Hyperparameters used in the DAE training and evaluation. The model was trained on a single Nvidia A40 GPU.}
\begin{tabular}{|l|l|}
\hline
\textbf{Hyperparameter}      & \textbf{Value}         \\ \hline
learning rate                & 0.0001                 \\ \hline
batch\_size                  & 64                     \\ \hline
image\_size                  & 96 x 96                \\ \hline
embedding\_size              & 512 \\ \hline
eval\_every\_samples         & 250,000                \\ \hline
precision                    & half                   \\ \hline
latent\_T\_eval              & 1,000                  \\ \hline
latent\_beta\_scheduler      & linear                 \\ \hline

total\_samples               & 12,000,000             \\ \hline
T                            & 1,000                  \\ \hline
eval\_T                      & 20                      \\ \hline
generation T                 & 100                    \\ \hline
\end{tabular}
\label{table:hyperparameters}
\end{table}

\end{document}